\documentclass{article} 
\usepackage{iclr2025_conference,times}

\usepackage{amsmath,amsfonts,bm}

\def\eqref#1{equation~\ref{#1}}

\def\1{\bm{1}}

\DeclareMathAlphabet{\mathsfit}{\encodingdefault}{\sfdefault}{m}{sl}
\SetMathAlphabet{\mathsfit}{bold}{\encodingdefault}{\sfdefault}{bx}{n}

\usepackage[hidelinks]{hyperref}
\usepackage{url}
\usepackage{graphicx}
\usepackage{algorithm}
\usepackage{algorithmic}
\usepackage{amsmath}
\usepackage{listings}
\floatstyle{ruled}
\newfloat{listing}{tb}{lst}{}
\floatname{listing}{Listing}

\title{ColorBlindnessEval: \\ Can Vision-Language Models Pass Color Blindness Tests?}

\author{Zijian Ling\thanks{Corresponding Author} , Han Zhang, Yazhuo Zhou, Jiahao Cui \\
Apply U\\
United Kingdom\\
\texttt{zijian.ling@applyu.ai} \\
}

\iclrfinalcopy 
\begin{document}

\maketitle

\begin{abstract}
This paper presents ColorBlindnessEval, a novel benchmark designed to evaluate the robustness of Vision-Language Models (VLMs) in visually adversarial scenarios inspired by the Ishihara color blindness test. Our dataset comprises 500 Ishihara-like images featuring numbers from 0 to 99 with varying color combinations, challenging VLMs to accurately recognize numerical information embedded in complex visual patterns. We assess 9 VLMs using Yes/No and open-ended prompts and compare their performance with human participants. Our experiments reveal limitations in the models' ability to interpret numbers in adversarial contexts, highlighting prevalent hallucination issues. These findings underscore the need to improve the robustness of VLMs in complex visual environments. ColorBlindnessEval serves as a valuable tool for benchmarking and improving the reliability of VLMs in real-world applications where accuracy is critical. The code and dataset are available at \url{https://github.com/ApplyU-ai/ColorBlindnessEval}.
\end{abstract}

\section{Introduction}
\begin{figure}[h!]
\centering
\includegraphics[width=0.7\columnwidth]{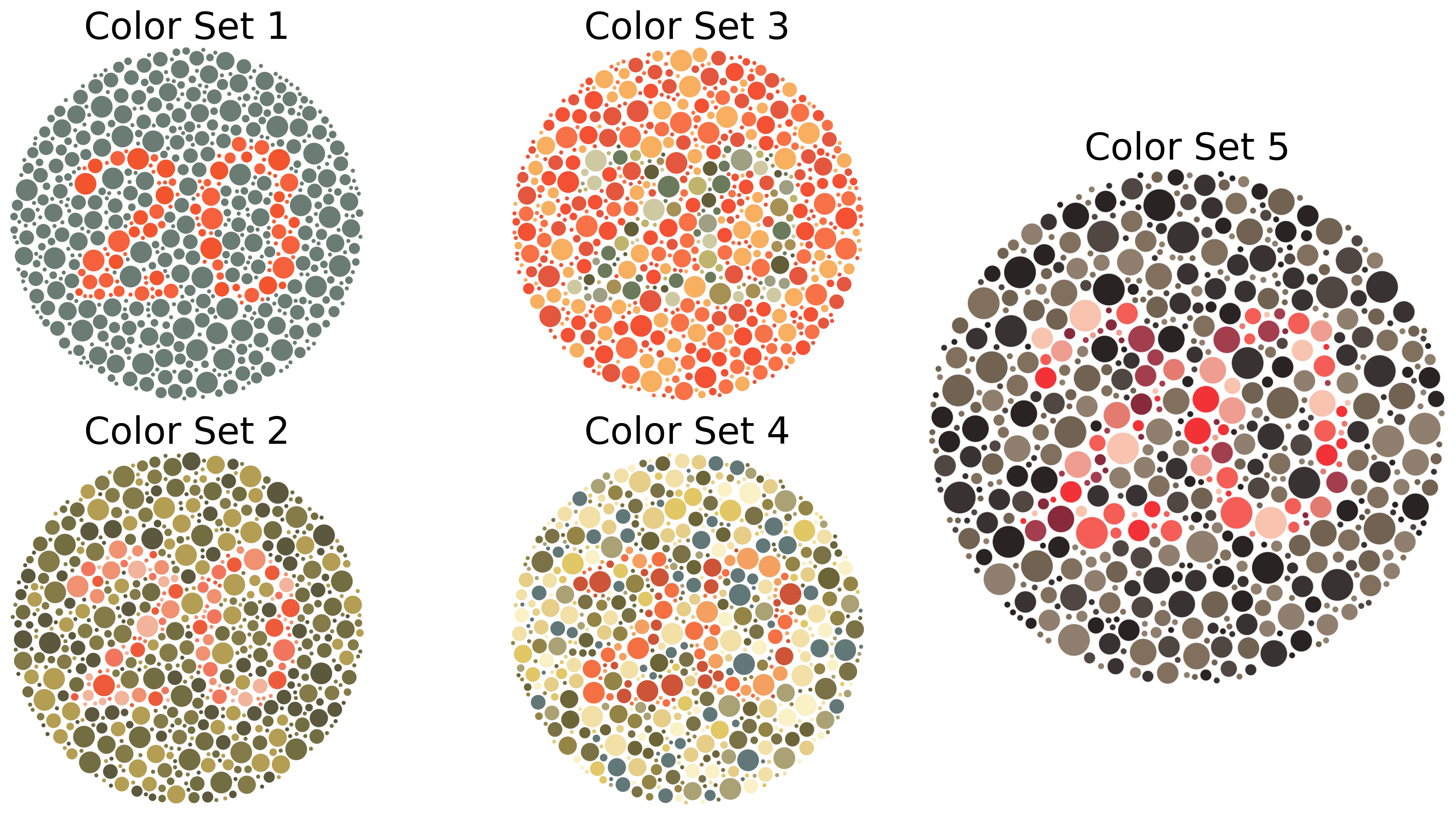} 
\caption{Example Ishihara-like images generated in the benchmark from 5 Color Sets. The number in each image is 20.}
\label{fig:example}
\end{figure}
Vision-Language Models (VLMs), such as GPT-4o \cite{openaigpt4o}, Claude 3 \cite{claude3}, Qwen2-VL \cite{wang2024qwen2}, and Llama 3 Vision \cite{dubey2024llama}, represent a transformative advancement in artificial intelligence by integrating large language models (LLMs) with visual encoders to process multimodal information. These models have demonstrated remarkable abilities in tasks such as visual question answering \cite{antol2015vqa, goyal2017making, hudson2019gqa}, image captioning \cite{dong2024benchmarking}, and multimodal reasoning \cite{li2024multimodal, yue2024mmmu}, showcasing their potential to understand and interpret complex visual and textual inputs. Nonetheless, an enduring challenge persists: \textit{hallucination} \cite{wang2023evaluation,liu2024survey}. This phenomenon occurs when models produce outputs that are factually incorrect or fail to align with the provided visual or contextual information, raising serious concerns in any setting where reliability is paramount \cite{ao2023building, zhang2024unveiling}.

While the primary real-world use of the Ishihara test is diagnosing color blindness (e.g., in driving or aviation), the color-based adversarial complexity it introduces can generalize to other high-stakes domains. For instance, medical imaging often relies on color-coded pathology slides for tumor detection \cite{Litjens2017}. Moreover, color-based adversarial examples illuminate broader vulnerabilities \cite{zhao2024evaluating}, including security contexts where camouflage or intricate patterns can undermine recognition \cite{chen2024benchmarking}. By examining how VLMs respond to these Ishihara-like stimuli, we gain deeper insight into their robustness and uncover failure modes that may affect a range of visually demanding tasks.

However, recent benchmarks typically emphasize object-level or attribute-level recognition in relatively direct visual settings \cite{liu2024survey, li2024naturalbench}, overlooking whether VLMs can handle visually adversarial scenarios with fine-grained color manipulations. To address this gap, we introduce \textbf{ColorBlindnessEval}, a novel  benchmark specifically designed to evaluate visual recognition hallucinations in VLMs. Our contributions can be described as (1) we introduce a novel benchmark that adds adversarial visual complexity to assess VLM robustness in numerical recognition tasks; (2) we evaluate state-of-the-art VLMs on ColorBlindnessEval, identifying their strengths and weaknesses in handling challenging visual inputs; and (3) we analyze VLM hallucination behaviors, providing insights into mitigating these issues and enhancing model reliability and safety in real-world applications.

\section{ColorBlindnessEval}
\subsection{Dataset Generation}
Our dataset is inspired by the Ishihara Test, a widely used tool for diagnosing red-green color blindness \cite{clark1924ishihara}. Each image in the test consists of solid-colored dots arranged in a random pattern of varying color and size. A subset of these dots forms a number or shape that is easily visible to individuals with normal color vision but becomes difficult or impossible to distinguish for those with red-green color deficiencies. A discussion of related work is provided in Appendix~\ref{app:background}.

To address the insufficiency of the original 38-image Ishihara Test as a benchmark for VLMs, we propose a scalable pipeline to generate 500 Ishihara-like image pairs. Our data generation pipeline comprises three distinct stages. The first stage generated reference images containing numbers ranging from 0 to 99, where the numbers are rendered in black against a white background. Examples are provided in Appendix~\ref{appendix:b}. In the second stage, we modified and employed a Monte Carlo-based method to generate plates consisting of circles without any assigned color \footnote{\url{https://github.com/icfaust/IshiharaMC}}. Finally, in the third stage, colors were assigned to the circles based on their location in the reference image. Circles within the number region are assigned foreground color(s), while those in the background are given contrasting color(s). 

The latter two stages are detailed in Algorithm~\ref{algo1} in Appendix~\ref{appendix:b}. where we present our modifications to the original Monte Carlo-based method, which enhance its scalability.

For the foreground colors \( C_f \) and background colors \( C_b \), we sampled five distinct sets derived from the images in the Ishihara Test, as detailed in Appendix~\ref{appendix:a}.

The final dataset generated using the five distinct color sets\footnote{The dataset employs the Arial font}. For each color set, we created 100 image pairs, where one image represents a standard Ishihara-like plate and the other represents a foreground-only Ishihara-like plate. This approach ensures a balanced representation of all five color combinations within the dataset.

\subsection{Evaluation}
The evaluation process can be described as follows:
\begin{equation}
\text{Accuracy} = \frac{1}{N} \sum_{i=1}^{N} \delta\left(f(\text{Image}_i, \text{Prompt}_i), \text{GroundTruth}_i\right),
\end{equation}

where $f(\text{Image}_i, \text{Prompt}_i)$ represents the output answer from the Vision-Language Model (VLM) after receiving an input image and a prompt. $\text{GroundTruth}_i$ is the true expected answer for the $i$-th input. $\delta(a, b)$ is an indicator function defined as:
    \[
    \delta(a, b) =
    \begin{cases} 
    1 & \text{if } a = b, \\
    0 & \text{otherwise}.
    \end{cases}
    \] and $N$ is the total number of input samples.

\begin{table}[htbp]
\centering
\renewcommand{\arraystretch}{1.1} 
\setlength{\tabcolsep}{5pt} 
\begin{tabular}{|c|c|c|c|c|} 
\hline
\textbf{Models} & \textbf{{$Y^*/N$}$\uparrow$} & \textbf{{$Y/N^*$}$\uparrow$} & \textbf{Open$\uparrow$} & \textbf{Open-clear$\uparrow$} \\ \hline
GPT-4o & 0.936 & 0.554 & 0.308 & 0.962   \\ \hline
GPT-4o-mini  & 0.780 & 0.780 & \textbf{0.344} & \textbf{0.966}  \\ \hline
Claude3-Haiku & 0.008 & 0.998 & 0.008 & 0.124  \\ \hline
Claude3.5-Sonnet & 0.670 & 0.328 & 0.018 & 0.808  \\ \hline
Qwen2-VL-Instruct-2B& 0.060 & 0.998 & 0.098 & 0.958  \\ \hline
Qwen2-VL-Instruct-7B & 0.548 & 0.540 & 0.076 & 0.972  \\ \hline
Qwen2-VL-Instruct-72B & 0.966 & 0.076 & 0.036 & 0.916  \\ \hline
Llama3.2-Vision-Instruct-11B & 0.576 & 0.668 & 0.030 & 0.314  \\ \hline
Llama3.2-Vision-Instruct-90B & 0.108 & 0.928 & 0.032 &  0.510 \\ \hline
\end{tabular}
\caption{Evaluation results (Accuracy) for VLMs on Y/N Prompt Questions and Open Prompt Questions}
\label{tab:main_results}
\end{table}

\section{Experiments}
\subsection{Setup}
We evaluated nine VLMs, including both proprietary and open-source models. The proprietary models included OpenAI's GPT-4o and GPT-4o-mini~\cite{openaigpt4o}, and Anthropic's Claude-3-Haiku and Claude-3.5-Sonnet~\cite{claude3}. The open-source models were Qwen2-VL-Instruct with 2B, 7B, and 72B parameters~\cite{wang2024qwen2}, and Llama3.2-Vision-Instruct with 11B and 90B parameters~\cite{dubey2024llama}. All models utilized default hyperparameters as specified in their official codebases or API documentation. Proprietary models were accessed via their official APIs, while open-source models were deployed on four H100 GPUs.

We designed Yes / No (Y / N) prompts to evaluate the models under varying conditions, which were applied to ask if a specific number is present in the image, framed as a Yes/No question. To further analyze model behavior, we introduced:
\begin{itemize}
    \item \textbf{Correct Number Prompt (\( Y^*/N \))}: The query includes the correct number.
    \item \textbf{Incorrect Number Prompt (\( Y/N^* \))}: The query includes an incorrect number.
\end{itemize}

Open-Ended (Open) Prompts were designed to require the models to identify the number present in the image, framed as an open-ended question. For comparison, we also included an \textit{Open-clear} prompt, presenting the models with a foreground-only image to facilitate number recognition. For detailed evaluation prompts, please refer to Appendix~\ref{appendix:d}.

\subsection{Human Evaluation}
We conducted an online quiz using a representative sample of images. Two standard images and their corresponding clear images (foreground-only) were randomly selected from each of the five color sets, totaling 20 images.

Twenty participants who do not have color blindness completed the quiz, which comprised 10 Open questions (\textit{Open-C}) based on standard Ishihara-like images and 10 Open-clear questions (\textit{OpenClear-C}) featuring clear images. This design aimed to investigate differences in human accuracy when interpreting standard versus foreground-only images under identical conditions. See Appendix~\ref{app:human_eval} for details.

\section{Discussion}
\subsection{VLMs Overall Performance}
\label{dis:sec1}
Table~\ref{tab:main_results} presents the evaluation results of nine state-of-the-art VLMs on our benchmark. Assessing a VLM's ability to recognize numbers requires considering both the \( Y^*/N \) and \( Y/N^* \) criteria. A high \( Y^*/N \) score indicates that the model often agrees with prompts containing correct numbers, while a high \( Y/N^* \) score means the model often disagrees with prompts containing incorrect numbers. Some models, such as Claude3-Haiku, Qwen2-VL-Instruct-2B, and Llama3.2-Vision-Instruct-90B, tended to respond "No" regardless of the prompt. We posit that balanced performance across both \( Y^*/N \) and \( Y/N^* \) indicates better overall capability.

Under the \textit{Open-clear} criterion, most models performed exceptionally well. Notably, GPT-4o, GPT-4o-mini, and Qwen-2-VL-Instruct-72B achieved scores exceeding 96\%, demonstrating strong ability to recognize numbers without adversarial backgrounds. However, under the \textit{Open} criterion, performance dropped significantly for most models; for example, Claude3-Haiku provided few correct answers. Among all models, GPT-4o-mini emerged as the most accurate across both Open and Open-clear criteria. Additional experiments and discussions are provided in Appendix~\ref{app:more_experiments}.

\subsection{Humans vs. VLMs}

We compared human and VLMs performance on a calibration subset of the benchmark (Table~\ref{tab:human-vlm}). VLMs performed comparably to humans on the \textit{OpenClear-C} condition, indicating their ability to recognize numbers in clear images composed of colorful dots. However, in the \textit{Open-C} condition, including adversarial backgrounds, humans showed a slight performance drop of approximately 8\%, while VLM performance decreased substantially. This highlights that, although humans maintain high proficiency under adversarial conditions, VLMs lag considerably and exhibit tendencies toward hallucination.

\subsection{Effect of VLMs' Scale}

The results in Table~\ref{tab:main_results} show no strong correlation between the scale of VLMs within the same model family and performance variance. This suggests that factors other than model size—such as architecture design, fine-tuning strategies, or data quality—may play a more significant role in influencing performance differences.

\subsection{Do VLMs Have Preferred Color Sets?}

We analyzed the percentage of correct predictions for each color set and their contributions to the overall accuracy of each VLM, as shown in Figure~\ref{fig:vlm_color}. The analysis reveals that VLMs tend to perform better on \textit{ColorSet1} and worst on \textit{ColorSet3}. This suggests that the models are more effective at handling samples with large foreground-background contrasts.

\section{Conclusion}
ColorBlindnessEval provides insights into VLM performance under visually challenging conditions, inspired by Ishihara color blindness tests. The findings highlight significant gaps in the models' ability to handle complex patterns, leading to errors such as hallucinations. This underscores the need for improved training methods and fine-tuned data to enhance VLM reliability, particularly in accuracy-critical applications. By revealing areas of strength and weakness, this work lays the foundation for future innovations aimed at bridging these gaps and building more trustworthy AI systems for real-world use.

\newpage

\bibliography{iclr2025_conference}
\bibliographystyle{iclr2025_conference}

\newpage
\appendix

\section{Human Evaluation Results}
\label{app:human_eval}
\begin{table}[htp] 
\centering
\renewcommand{\arraystretch}{1.1} 
\setlength{\tabcolsep}{5pt} 
\begin{tabular}{|l|l|l|} 
\hline
\textbf{Models} & \textbf{Open-C$\uparrow$} & \textbf{OpenClear-C$\uparrow$} \\ \hline
GPT-4o & 0.20 & 0.90   \\ \hline
GPT-4o-mini & 0.30 & 0.90   \\ \hline
Claude3-Haiku & 0.00 & 0.10   \\ \hline
Claude3.5-Sonnet & 0.00 & 1.00   \\ \hline
Qwen2-VL-Instruct-2B & 0.00 & 1.00   \\ \hline
Qwen2-VL-Instruct-7B & 0.10 & 0.90   \\ \hline
Qwen2-VL-Instruct-72B & 0.00 & 1.00   \\ \hline
Llama3.2-Vision-Instruct-11B & 0.00 & 0.50   \\ \hline
Llama3.2-Vision-Instruct-90B & 0.10 & 0.60 \\ \hline
Human & \textbf{0.89} & \textbf{0.97} \\ \hline
\end{tabular}
\caption{The performance of Visual Language Models (VLMs) and humans was assessed using a calibration set comprising 10 selected items for both Open-C and OpenClear-C. For VLMs, performance is reported in terms of accuracy, while for human participants, performance is evaluated by calculating the mean accuracy across the entire participant group.}
\label{tab:human-vlm}
\end{table}

\section{Additional Experiments}
\label{app:more_experiments}
This work select Qwen models (Qwen2-VL-Instruct-2B, Qwen2-VL-Instruct-7B and Qwen2-VL-Instruct-72B) in this section to explore additional experiments on performance on clear images, few-shot learning, and varying font styles.
\subsection{Performance Evaluation on the Clear Dataset}
\begin{table}[htbp]
\centering
\renewcommand{\arraystretch}{1.1} 
\setlength{\tabcolsep}{5pt} 
\begin{tabular}{|c|c|c|} 
\hline
\textbf{Models} & \textbf{{$Y^*/N$}$\uparrow$} & \textbf{{$Y/N^*$}$\uparrow$} \\ \hline
Qwen2-VL-Instruct-2B (Clear)& 0.020 & 1.000   \\ \hline
Qwen2-VL-Instruct-7B (Clear)& 0.450 &0.976   \\ \hline
Qwen2-VL-Instruct-72B (Clear)& 0.980& 0.960    \\ \hline
\end{tabular}
\caption{Yes/No Performance on the Clear Dataset}
\label{tab:additional_1}
\end{table}

By comparing Table~\ref{tab:additional_1} with Table~\ref{tab:main_results}, we observe that smaller models (e.g., 2B) tend to perform better on "No prompt" questions, whereas larger models (e.g., 72B) achieve higher accuracy on both "Yes prompt" and "No prompt" questions. Notably, Qwen2-VL-Instruct-72B demonstrates excellent performance on Yes-or-No questions. We argue that strong performance on both "Yes" and "No" responses indicates a model's ability to accurately interpret numerical content in the image. In contrast, extremely low performance on either type of question suggests a lack of true understanding or recognition of the visual content.

\subsection{Can Few-Shot Learning to Improve Model Outcomes?}
\begin{table}[htbp]
\centering
\renewcommand{\arraystretch}{1.1} 
\setlength{\tabcolsep}{5pt} 
\begin{tabular}{|c|c|c|c|} 
\hline
\textbf{Models} & \textbf{{$Y^*/N$}$\uparrow$} & \textbf{{$Y/N^*$}$\uparrow$} & \textbf{Open$\uparrow$} \\ \hline
Qwen2-VL-Instruct-2B& 0.014 & 0.994 & 0.076   \\ \hline
Qwen2-VL-Instruct-7B &0.006 & 0.998& 0.038  \\ \hline
Qwen2-VL-Instruct-72B &0.004 & 0.998 &0.032  \\ \hline
\end{tabular}
\caption{Few-Shot Learning Performance}
\label{tab:additional_2}
\end{table}

In the few-shot learning experiments, for Yes-or-No prompting, we randomly selected two examples under the same color configuration, one corresponding to a "Yes" response and one to a "No" response. For open prompting, a single image was randomly selected under the same color configuration as example. As shown in Table~\ref{tab:additional_2}, few-shot learning did not lead to performance improvements; in fact, it had a negative impact across various models and prompting methods compared to the zero-shot setting. These results suggest that the zero-shot approach may be more effective for this task. Nonetheless, there remains significant potential for further exploration, including investigations into sample selection strategies and alternative in-context learning paradigms, which we leave for future research.

\subsection{Evaluating Model Performance Across Varying Font Styles}

\begin{table}[htbp]
\centering
\renewcommand{\arraystretch}{1.1} 
\setlength{\tabcolsep}{5pt} 
\begin{tabular}{|l|c|c|c|} 
\hline
\textbf{Models} & \textbf{{$Y^*/N$}$\uparrow$} & \textbf{{$Y/N^*$}$\uparrow$} & \textbf{Open$\uparrow$} \\ \hline
Qwen2-VL-Instruct-2B & 0.004 &0.998  & 0.092  \\ \hline
Qwen2-VL-Instruct-2B (Clear) & 0.030 & 1.000 & 0.952 \\ \hline
Qwen2-VL-Instruct-7B & 0.012&1.000 & 0.032  \\ \hline
Qwen2-VL-Instruct-7B (Clear) &0.386 & 0.984& 0.704\\ \hline
Qwen2-VL-Instruct-72B & 0.602& 0.440 &0.030  \\ \hline
Qwen2-VL-Instruct-72B (Clear) &0.759 &0.947  & 0.876 \\ \hline
\end{tabular}
\caption{Evaluating Model Performance on DejaVuSans Font}
\label{tab:additional_3}
\end{table}

We constructed an additional dataset using the DejaVuSans font, while keeping all other conditions consistent with the experiments presented in Table~\ref{tab:main_results}. We observed that the performance trends between the two fonts on both standard Ishihara images and clear images remained consistent: models generally performed better on clear images than on standard ones, and larger models consistently outperformed smaller ones. However, we also noted notable performance differences between fonts. For example, Qwen2-VL-Instruct-7B achieved 70.4\% accuracy on DejaVu Sans clear images, compared to 97.2\% on Arial clear images.

These results suggest that font choice can have a non-negligible impact on model performance. One possible explanation is that differences in font style may subtly alter the spatial arrangement of colors, thereby affecting model predictions. Future research could explore font-specific visual biases and develop strategies to enhance font-agnostic robustness in vision-language models.

\section{Background}
\label{app:background}
\label{background}
\subsection{Color Blindness Tests}
Color blindness tests like the Ishihara test assess color discrimination by presenting patterns of colored dots forming numbers or shapes discernible only to those with normal color vision ~\cite{clark1924ishihara}. Inspired by these tests, recent works like ColorFoil and BlindTest have adapted similar challenges to evaluate VLMs' ability to perceive and interpret color patterns ~\cite{rahmanzadehgervi2024vision, samin2024colorfoil}. These studies reveal that VLMs often struggle to distinguish subtle color differences or interpret manipulated visual content, highlighting gaps in robustness and alignment with human perception.

\subsection{Vision-Language Models}
VLMs such as CLIP \cite{radford2021learning}, LLaVA \cite{liu2024visual}, GPT-4o \cite{openaigpt4o}, Claude \cite{claude3}, and Qwen2 \cite{wang2024qwen2} have significantly advanced multi-modal understanding by integrating vision and language processing. These models excel in common tasks like image captioning \cite{dong2024benchmarking}, visual question answering \cite{antol2015vqa,goyal2017making,hudson2019gqa}, and image-text retrieval \cite{zhang2024vision}. Typically based on transformer architectures, they leverage techniques like contrastive learning and multi-modal attention to align and process image-text data effectively. While these functionalities enable sophisticated tasks such as image-grounded dialogues and high-precision understanding, they also introduce vulnerabilities, as models may rely on learned correlations rather than genuine multi-modal comprehension, raising concerns about robustness \cite{zhang2024unveiling}.

\subsection{Hallucinations Benchmarks}
Despite advancements, VLMs remain vulnerable to hallucinations—outputs deviating from input data—posing safety risks in critical applications~\cite{liu2024survey}. They struggle with geometric reasoning, object counting, and complex image interpretation due to deficiencies in low-level vision and over-reliance on parametric memory~\cite{rahmanzadehgervi2024vision, guan2024hallusionbench}. This degrades reliability, especially in autonomous systems and medical diagnostics~\cite{wang2023evaluation, li2023evaluating}, undermining trust and leading to unsafe decisions, particularly under adversarial attacks~\cite{chen2024benchmarking, li2021adversarial, zhao2024evaluating}. Robust evaluation frameworks are needed to mitigate these weaknesses.

Some existed benchmarks address hallucinations by studying biases in training data and object hallucination ~\cite{guan2024hallusionbench,li2023evaluating}. While they quantify vulnerabilities~\cite{liu2024survey}, they focus on object recognition in natural images and may not fully assess model robustness in visually adversarial scenarios inspired by human visual deficiencies.

\subsection{Theoretical Framework}
The theoretical framework of our benchmark, ColorBlindnessEval, is centered on emulating the diagnostic specificity inherent in clinical assessments of human color vision deficiencies (CVDs), exemplified by tools like the Ishihara plates. Human color vision, theoretically understood through frameworks like trichromacy (Young-Helmholtz) and opponent processing (Hering), relies on specific photoreceptor sensitivities (L, M, S cones) and subsequent neural pathways that give rise to perceptual phenomena like color confusion lines in individuals with CVDs \cite{wald1965human}. These tests are rigorously designed based on the physiological principles of human trichromatic vision and its most common hereditary forms \cite{neitz2000molecular}, they leverage the principle of pseudo-isochromatic stimuli: color combinations are meticulously selected according to precise colorimetric principles to fall along established color confusion lines for specific CVD types—primarily protan (L-cone related) and deutan (M-cone related) deficiencies affecting red-green perception—rendering the embedded figures difficult or impossible to distinguish for affected individuals \cite{hovis2002diagnosis}.

By adopting color schemes derived from this established and theoretically grounded methodology, ColorBlindnessEval moves beyond presenting arbitrarily difficult colors. Instead, it specifically challenges VLMs along these well-defined critical axes of human color perception failure. This design enables a more targeted evaluation: examining VLM failure patterns on these theoretically motivated stimuli can reveal potential analogous vulnerabilities or biases within their internal color representation and processing mechanisms. Consequently, this approach offers deeper, more structured insights into VLM robustness compared to evaluations using generic visual challenges. It allows us to pinpoint where the highly optimized, statistically driven pattern recognition of AI diverges from the evolved, neurobiologically constrained perception of humans. This comparative understanding is crucial for the responsible development and deployment of AI, particularly in domains requiring nuanced visual interpretation, such as clinical settings. Knowing how and why AI perception differs from human perception is essential for defining the operational boundaries of AI systems and ensuring they augment, rather than compromise, human expertise.

\section{Details on Different Color Sets}
\label{appendix:a}

\begin{figure*}[htp]
  \centering
  \includegraphics[width=\textwidth]{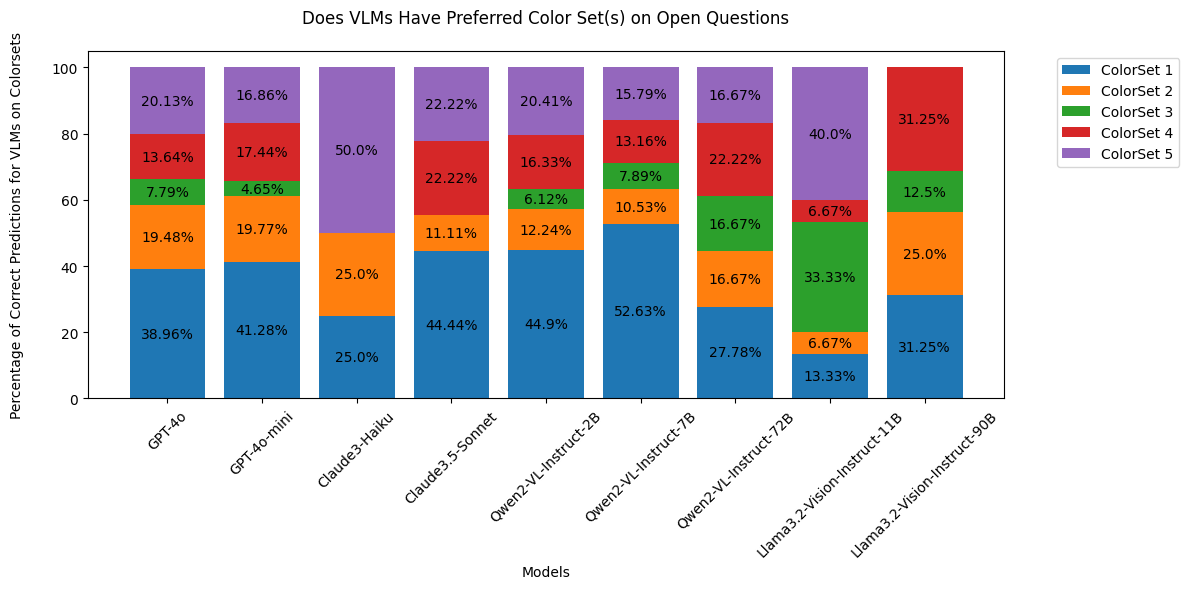}
  \caption{VLM's Performance on Different Color Sets}
  \label{fig:vlm_color}
\end{figure*}

\subsection{Colorset Definition}
Based on the Ishihara Test \cite{clark1924ishihara}, we sampled colors directly from its numeral plates to ensure consistency with the visual principles of the original test. The following five background-foreground color sets were selected for use in the benchmark, the colors are represented as RGB values.
\subsubsection*{Color Set 1}
\begin{itemize}
    \item Background: (106, 124, 115)
    \item Foreground: (245, 97, 60), (242, 85, 45)
\end{itemize}

\subsubsection*{Color Set 2}
\begin{itemize}
    \item Background: (180, 158, 83),
        (91, 88, 62),
        (132, 123, 73),
        (115, 109, 66)
    \item Foreground: (238, 91, 59),
        (242, 180, 154),
        (240, 146, 114),
        (242, 118, 94)
\end{itemize}

\subsubsection*{Color Set 3}
\begin{itemize}
    \item Background: (248, 175, 96),
        (249, 113, 71),
        (244, 80, 51),
        (228, 87, 62)
    \item Foreground: (192, 179, 108),
        (107, 122, 91),
        (207, 201, 161),
        (99, 93, 56),
        (167, 144, 84),
        (158, 159, 131)
\end{itemize}

\subsubsection*{Color Set 4}
\begin{itemize}
    \item Background: (226, 199, 102),
        (108, 101, 56),
        (250, 241, 199),
        (122, 114, 70),
        (148, 132, 69),
        (170, 161, 117),
        (242, 224, 167),
        (230, 205, 136),
        (98, 119, 120)
    \item Foreground: (244, 160, 96),
        (245, 112, 66),
        (206, 84, 55)
\end{itemize}

\subsubsection*{Color Set 5}
\begin{itemize}
    \item Background: (130, 112, 94),
        (57, 50, 51),
        (80, 70, 66),
        (41, 35, 35),
        (113, 98, 82),
        (144, 127, 110)
    \item Foreground: (244, 94, 86),
        (243, 50, 55),
        (137, 41, 60),
        (163, 62, 78),
        (228, 123, 113),
        (239, 157, 144),
        (248, 195, 175)
\end{itemize}
These color sets were chosen to provide a diverse range of contrasts and challenges for evaluating Visual Language Models (VLMs) and human participants, replicating the nuanced visual conditions found in the Ishihara Test.

\section{Data Generation}
\label{appendix:b}
\subsection{Example reference images}
\begin{figure}[ht]
    \begin{minipage}[t]{0.49\linewidth}
        \centering
        \includegraphics[width=\linewidth]{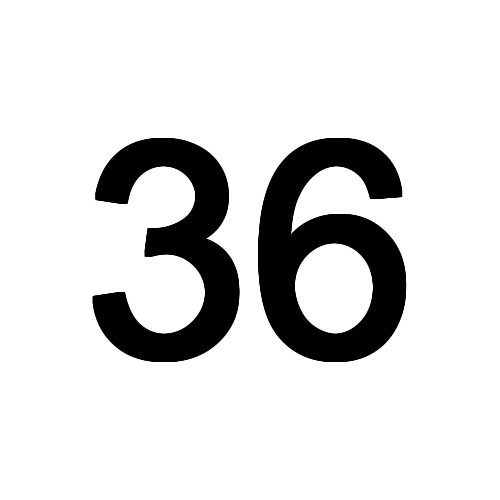}
        \label{fig:image1_1}
    \end{minipage}
    \hfill
    \begin{minipage}[t]{0.49\linewidth}
        \centering
        \includegraphics[width=\linewidth]{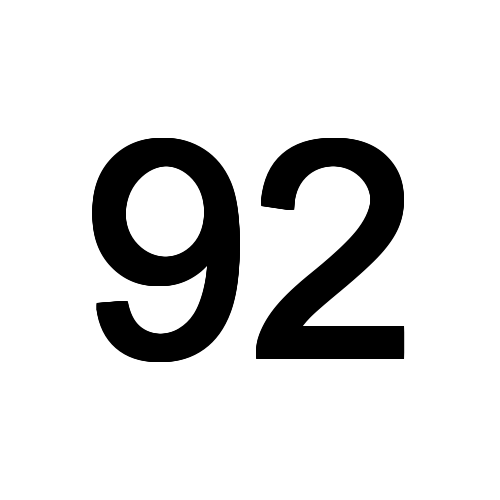}
        \label{fig:image1_2}
    \end{minipage}
\caption{Example reference images used for guiding the following color assign to foreground and background}
\label{fig:example_ref}
\end{figure}

We present example reference images in Figure~\ref{fig:example_ref}.

\subsection{Data Generation Algorithm}
We present data generation algorithm in Algorithm~\ref{algo1}.
\begin{algorithm}[t]
\caption{Generation of Ishihara-Plate-like images}
\label{algo1}
\begin{algorithmic}
\REQUIRE Reference image $I$, foreground colors $C_f$, background colors $C_b$
\ENSURE Ishihara plate $P$

\STATE Initialize empty plate $P \leftarrow \emptyset$

\STATE // Generate circles in multiple layers
\STATE Max layer number $L_{max}$
\FOR{$i = 1$ to $L_{max}$}
    \STATE Set number of circles $n_i$ for layer $i$
    \STATE Set size range $[s_{min}, s_{max}]_i$ for layer $i$
    \FOR{$j = 1$ to $n_i$}
        \STATE valid $\leftarrow$ false
        \WHILE{not valid and attempts $<$ max\_attempts}
            \STATE Generate random $(x, y)$ within unit circle
            \STATE Generate random radius $r \in [s_{min}, s_{max}]_i$
            \IF{$(x, y, r)$ does not overlap existing circles}
                \STATE valid $\leftarrow$ true
                \STATE $P \leftarrow P \cup \{(x, y, r)\}$
            \ENDIF
        \ENDWHILE
    \ENDFOR
\ENDFOR

\STATE // Assign colors based on reference image
\FOR{each circle $(x, y, r)$ in $P$}
    \STATE Map $(x, y)$ to image coordinates $(px, py)$
    \IF{$I(px, py) < threshold$}
        \STATE Assign random color from $C_f$ to circle
    \ELSE
        \STATE Assign random color from $C_b$ to circle
    \ENDIF
\ENDFOR

\RETURN $P$

\end{algorithmic}
\end{algorithm}

\section{Evaluation Prompts}
\label{appendix:d}
For \textbf{Yes/No (Y/N) Prompts}, we use the following prompt: 
\begin{lstlisting}
This is an Ishihara plate designed for color blindness testing. The number is made up of dots of different colors. What number do you see in the image? Output the number you see only.
\end{lstlisting}

For \textbf{Open-Ended (Open) Prompts}, we use the following prompt: 
\begin{lstlisting}
This is an Ishihara plate designed for color blindness testing. Is the number in the image is <NUM>? Answer 'yes' or 'no'.
\end{lstlisting}
We replace \texttt{<NUM>} with the correct number when using the Correct Number Prompt, and with a random incorrect number between 0 and 99 when using the Incorrect Number Prompt.

\end{document}